\documentclass{article}
\usepackage{iclr2027_conference,times}
\iclrfinalcopy
\usepackage{iftex}
\ifPDFTeX
  \usepackage[utf8]{inputenc} 
\fi
\usepackage[T1]{fontenc}    
\usepackage{hyperref}       
\usepackage{url}            
\usepackage{booktabs}       
\usepackage{amsfonts}       
\usepackage{nicefrac}       
\usepackage{microtype}      
\usepackage{xcolor}         
\usepackage{pifont}         
\usepackage{booktabs, multirow, tabularx} 
\usepackage{colortbl} 
\usepackage{graphicx}
\usepackage{caption}
\usepackage{arydshln}
\usepackage{amsmath}
\usepackage{cleveref}
\usepackage{subcaption}
\usepackage{tikz}
\usetikzlibrary{arrows.meta,positioning,fit,backgrounds,calc}
\usepackage{xspace}  

\title{VGGA: Geometry-Grounded Autoencoder for Video Generation}

\author{%
  Chaofan Gan\textsuperscript{1,2,*} \quad Yian Zhao\textsuperscript{3,*} \quad Shijie Li\textsuperscript{1} \quad 
  Tieyuan Chen\textsuperscript{1} \quad Zicheng Zhao\textsuperscript{1} \\ \textbf{Mehrtash Harandi\textsuperscript{2} \quad Weiyao Lin\textsuperscript{1,\textdagger}}\\[0.5em]
  \textsuperscript{1}Shanghai Jiao Tong University \quad 
  \textsuperscript{2}Monash University \quad \textsuperscript{3}Peking University
}

\begin{document}

\maketitle
\begingroup
  \renewcommand{\thefootnote}{\fnsymbol{footnote}}
  \footnotetext[1]{Equal contribution.}
  \footnotetext[2]{Corresponding author.}
\endgroup

\begin{figure}[htp]
  \centering
  \vspace{-10mm}
  \includegraphics[width=\linewidth]{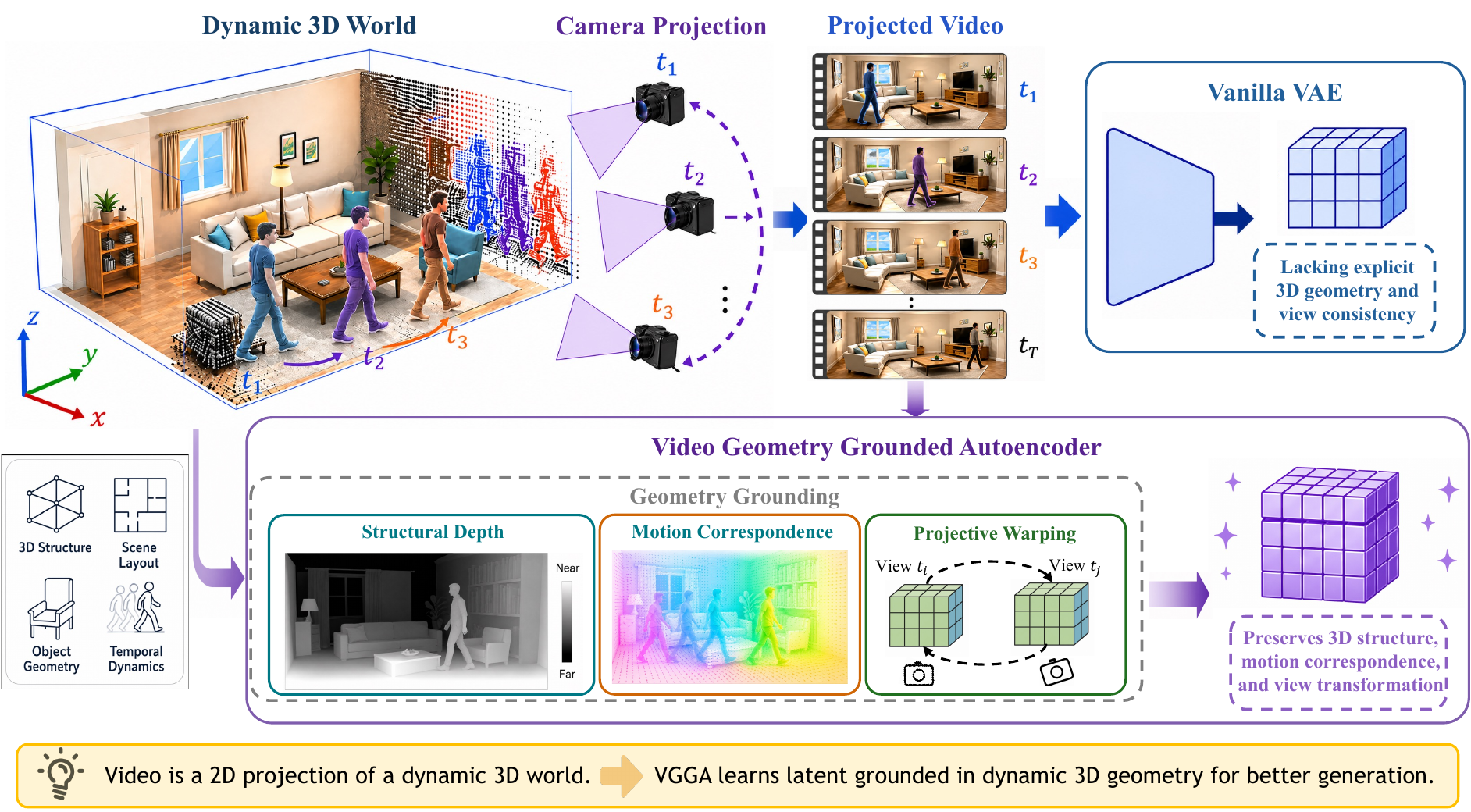}
  \vspace{-0.8em}
  \caption{\textbf{Geometry-grounded video autoencoding.}
Videos arise from the 2D projection of dynamic 3D worlds, whose underlying
geometry naturally contains scene layout, dense motion correspondence, and
camera-induced projective transformations. VGGA exploits these geometric regularities to organize video latents
beyond reconstruction-only supervision, yielding a more compact, structured,
and geometry-aware latent space for video generation.
}
  \label{fig:vgga_motivation}
  \vspace{-0.8em}
\end{figure}

\begin{abstract}
Autoencoders are the core component of latent video generative models, compressing high-dimensional videos into 
compact latent spaces for downstream generation. However, existing video autoencoders are primarily optimized 
for pixel-level reconstruction, providing limited supervision for learning structured latent representations. 
We argue that videos should not be treated merely as signals to be reconstructed, but rather as 2D projections 
of dynamic 3D worlds that exhibit explicit and well-structured geometric properties.
Building on this insight, we propose \textbf{V}ideo \textbf{G}eometry \textbf{G}rounded 
\textbf{A}utoencoder (\textbf{VGGA}), a geometry-aware video 
autoencoder that explicitly incorporates geometric constraints into the reconstruction process. 
Specifically, VGGA grounds the learned representations along three complementary geometric dimensions: 
\emph{Structural Depth}, \emph{Motion Correspondence}, and \emph{Projective Warping}. 
These objectives are unified within a Video Geometry Decoder, which employs a shared spatio-temporal 
upsampling trunk to jointly optimize geometric learning and RGB reconstruction. Through these geometry-grounded 
objectives, VGGA encourages the latent representations to capture the underlying 3D structure and motion 
of dynamic scenes.
VGGA consistently improves video generation across both geometry-centric and human-motion domains, 
reducing gFVD by 28.1\% on RealEstate10K and 37.7\% on Taichi-HD.
Notably, these substantial generation improvements are achieved while maintaining comparable reconstruction quality, 
demonstrating that geometry-grounded representation learning yields a more effective latent space for video generation.
\end{abstract}

\section{Introduction}

Latent diffusion models~\citep{seedance2026seedance,gao2025seedance,
wan2025wan,yang2024cogvideox} have become a dominant paradigm for
high-fidelity image and video generation. They typically employ a
Variational AutoEncoder (VAE)~\citep{kingma2013auto} to compress visual
signals into a compact latent space, over which a diffusion transformer
performs generative modeling~\citep{rombach2022high,peebles2023scalable,
ma2024latte,lin2024opensora,yu2024language}. Beyond compression, the VAE
determines what information is preserved and how the latent space is
structured, thereby substantially affecting downstream optimization and
generation quality~\citep{yao2025reconstruction,
skorokhodov2025improving}.

Despite its importance, most video autoencoders are still trained
primarily for pixel-level reconstruction, providing limited guidance for
learning generation-friendly latent representations. Recent works
address this issue by aligning VAE latents with vision foundation
models~\citep{yu2025representation,ma2025unitok,leng2025repae},
incorporating self-supervised representation learning~
\citep{ma2025unitok,yao2025towards}, or imposing latent regularization
such as smoothness and scale equivariance~
\citep{liu2025delving,skorokhodov2025improving,kouzelis2025eq}.
These approaches improve semantic structure and latent diffusability
beyond reconstruction alone.

However, these efforts still leave two limitations for video autoencoders. First,
understanding-oriented supervision often encourages invariance to
instance-specific details~
\citep{achille2018emergence,tian2020makes,li2023addressing}, whereas
autoencoding requires preserving fine-grained appearance and motion for
faithful reconstruction~\citep{yao2025reconstruction}, creating an
inherent objective mismatch~\citep{shi2025latent} (see \Cref{tab:kinetics400_reconstruction}). Second, existing
objectives provide little video-specific grounding. Most are inherited
from image representation learning or generic latent regularization,
while video arises from the projection of a dynamic 3D world, as
illustrated in \Cref{fig:vgga_motivation}. Without explicitly modeling
this underlying geometry, downstream generators must infer depth,
motion, and viewpoint consistency implicitly from compressed RGB
latents.

Therefore, we propose \textbf{V}ideo \textbf{G}eometry \textbf{G}rounded 
\textbf{A}utoencoder (\textbf{VGGA}),  
a geometry-aware video autoencoder for learning generation-friendly video latents.  
Concretely, we introduce a Video Geometry Decoder
that grounds the learned representations along three complementary geometric dimensions:  
\emph{Structural Depth Grounding} for per-frame 3D structure,  
\emph{Motion Correspondence Grounding} for dense temporal correspondence,  
and \emph{Projective Warping Grounding} for camera-induced view transformations.  
To build an effective VGGA, we focus on two key challenges:  
\textbf{(1) how to harmonize geometric learning with RGB reconstruction},  
allowing geometric supervision to enrich the latent representations  
without degrading reconstruction quality; and  
\textbf{(2) how to enable geometric grounding on large-scale videos}  
without explicit geometric annotations.

For the first challenge, a key difficulty is that video latents are compressed in both space and time,  
whereas geometric supervision is defined at the frame level.  
VGGA therefore adopts a partially shared decoder with a shared spatio-temporal upsampling trunk  
before the RGB reconstruction and geometry branches (see \Cref{fig:vgga_pipeline}).  
This design allows geometric supervisions to act directly on the decoded frame-level representations used for reconstruction,  
encouraging the latent space to encode geometric structure without sacrificing visual fidelity.

For the second challenge, we adopt a hybrid geometric supervision strategy.  
For general videos, we use pseudo depth and motion cues predicted by off-the-shelf models  
to perform \emph{Structural Depth Grounding} and \emph{Motion Correspondence Grounding}.  
In contrast, \emph{Projective Warping Grounding} requires reliable camera poses
and assumes static scene geometry, so it cannot account for independent object motion.
We therefore apply this objective only to a small set of predominantly static-scene videos with known camera poses.
Through joint training, the projective geometry learned from these static scenes  
is absorbed into the shared latent representation and complements the depth and motion supervision from general videos,  
allowing camera-aware geometric priors to transfer beyond the statically supervised data  
and improve generation on dynamic videos. Our main contributions are
summarized as follows.

\begin{itemize}

\item We present a new perspective on video autoencoding  
by viewing videos as 2D projections of dynamic 3D scenes,  
and propose \textbf{V}ideo \textbf{G}eometry \textbf{G}rounded 
\textbf{A}utoencoder (\textbf{VGGA})  
to inject explicit geometric priors into video latent learning.

\item We develop an effective geometry-grounded  
autoencoding framework that harmonizes geometric learning  
with RGB reconstruction and supports large-scale training  
without explicit geometric annotations.

\item We demonstrate that VGGA substantially improves downstream  
video generation while preserving reconstruction quality,  
demonstrating the effectiveness and compatibility of  
geometric priors for video autoencoding.

\end{itemize}

\section{Related Work}

\noindent\textbf{Video Autoencoders for Latent Generation.}
Latent generative models~\citep{seedance2026seedance, lin2024opensora} hinge on an underlying autoencoder to define the space for generative modeling. While Latent Diffusion Models (LDMs)~\citep{rombach2022high} demonstrated that operating in a compressed latent space reconciles computational efficiency with visual fidelity, Diffusion Transformers (DiTs)~\citep{peebles2023scalable, ma2024sit} further validated the scalability of this paradigm. This design is particularly pivotal for video, where the tokenizer must encapsulate both spatial appearance and temporal dynamics~\citep{gao2025seedance, yang2024cogvideox, wan2025wan}. Despite the heavy reliance of current systems on autoencoder quality~\citep{liu2025delving}, the training objective remains predominantly reconstruction-centric. 

\noindent\textbf{Generation-Friendly Tokenizer Learning.}
Recent studies~\citep{yao2025reconstruction} suggest that reconstruction fidelity alone does not guarantee effective generative latents, sparking interest in generation-centric tokenization~\citep{ma2025unitok, yao2025towards, skorokhodov2025improving}. One research trajectory aligns VAE latents with vision foundation models (e.g., DINOv2) to infuse semantic and spatial priors~\citep{yu2025representation, leng2025repae}. Others integrate understanding-oriented objectives, such as contrastive or self-supervised learning, to refine latent organization~\citep{ma2025unitok, yao2025towards}. Parallel efforts focus on latent "diffusability" through regularization---leveraging spectral biasing or equivariance~\citep{liu2025delving, skorokhodov2025improving, kouzelis2025eq}. While these methods enhance latent structure, they primarily adapt image-level paradigms or generic constraints, largely overlooking the intrinsic priors of video formation. VGGA fills this gap by explicitly grounding video autoencoding in geometric constraints.

\noindent\textbf{Geometry Grounding for Video as Dynamic 3D Worlds.}
Videos are inherently projections of dynamic 3D scenes, positioning geometry as a fundamental source of dense, self-contained supervision. While extensive research has focused on individual geometric cues, such as monocular depth~\citep{yang2024depth}, optical flow~\citep{teed2020raft}, and point tracking~\citep{karaev2024cotracker}, or leveraged projective consistency for novel view synthesis~\citep{wang2024dust3r}, these efforts primarily target geometric perception. These works validate that geometry effectively distills the physical regularities manifest in video observations. VGGA departs from these perception-centric goals by repurposing geometric cues as grounding signals for video autoencoding. Consequently, the resulting latent space is explicitly sculpted by 3D layout and temporal correspondence, moving beyond the limitations of purely pixel-based reconstruction.

\section{Method}
\begin{figure}[!t]
  \centering
  \includegraphics[width=\linewidth]{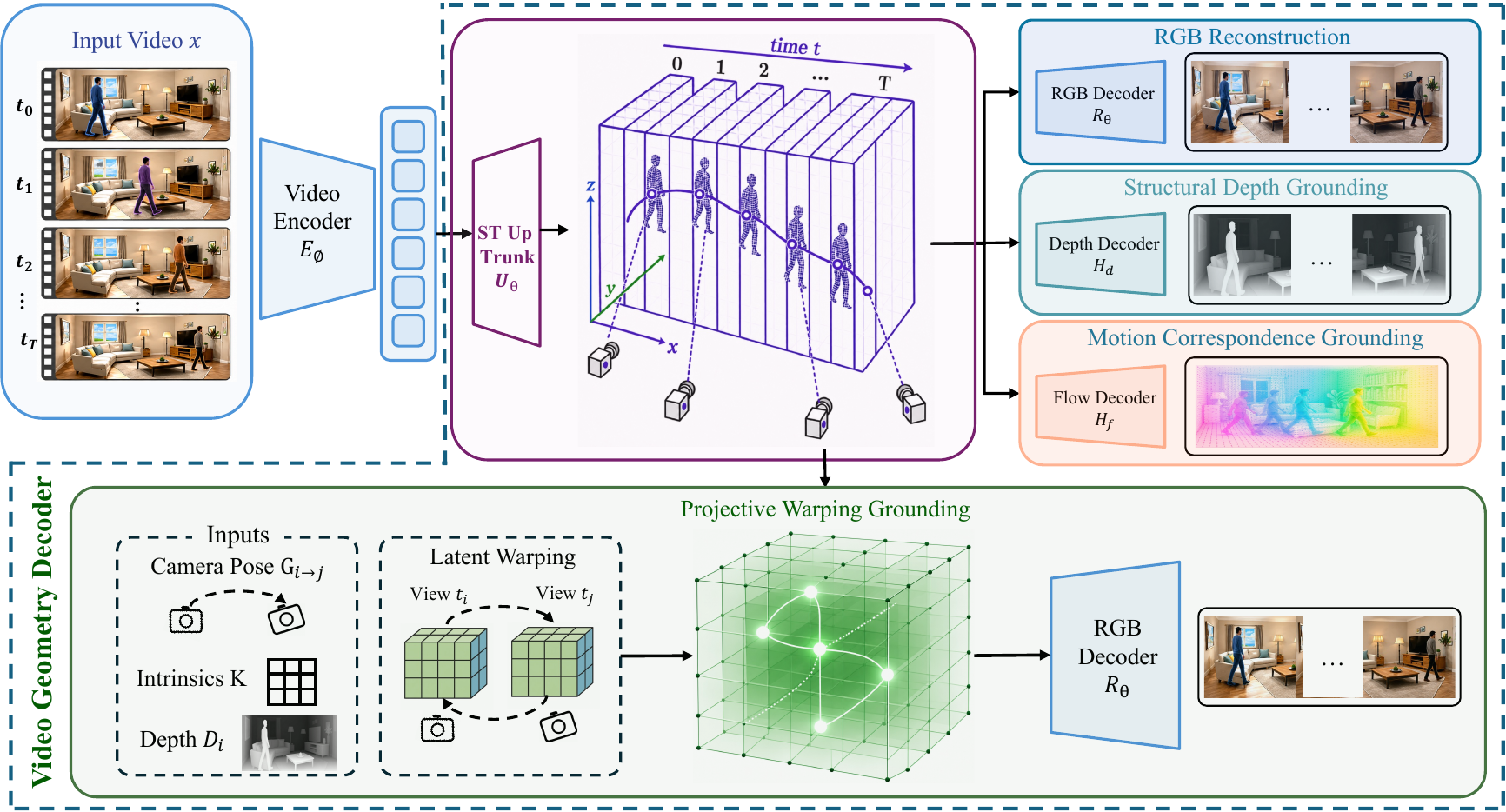}
  \vspace{-0.8em}
  \caption{\textbf{Overview of VGGA.} Our VGGA introduces a Video Geometry Decoder to ground
  compressed video latents in dynamic 3D geometry. A shared Spatio-Temporal
  Upsampling Trunk (ST UP Trunk) lifts the latent to a frame-aligned decoded representation,
  on which RGB reconstruction jointly acts with three geometry grounding tasks,
  including structural depth, motion correspondence, and projective warping.}
  \label{fig:vgga_pipeline}
  \vspace{-1.0em}
\end{figure}

We propose VGGA to ground video autoencoding in the geometry of dynamic 3D
worlds. As shown in \Cref{fig:vgga_pipeline}, VGGA replaces a purely
reconstruction-driven decoder with a Video Geometry Decoder. The decoder
builds a shared decoded representation for RGB reconstruction and three
geometry grounding signals: structural depth, motion correspondence, and
projective warping. We start with preliminaries in \Cref{sec:preliminary}, followed by the Video
Geometry Decoder in \Cref{sec:video_geometry_decoder}, Video Geometry Grounding in \Cref{sec:geometry_grounding}, Hybrid Geometry Supervision on Unlabelled Videos in \Cref{sec:scalable_grounding}.

\subsection{Preliminary}
\label{sec:preliminary}
We first introduce the notation and basic objectives used throughout the paper.
Let $\mathbf{x}=\{I_t\}_{t=0}^{T}\in\mathbb{R}^{(1+T)\times H\times W\times 3}$
denote a video clip with $1+T$ RGB frames, where $I_t$ is the frame at time $t$.

\noindent\textbf{Video Latent Autoencoding.}
A video VAE~\citep{kingma2014auto} compresses an input video
$\mathbf{x}\in\mathbb{R}^{(1+T)\times H\times W\times 3}$ into a compact
spatio-temporal latent
$\mathbf{z}\in\mathbb{R}^{(1+T/r_t)\times H/r_s\times W/r_s\times C}$, where
$r_t$ and $r_s$ denote the temporal and spatial downsampling ratios and $C$ is
the latent channel dimension. The decoder then reconstructs the latent back to
RGB space. Specifically, the encoder $E_{\phi}$ parameterizes a posterior
distribution
\begin{equation}
q_{\phi}(\mathbf{z}\mid \mathbf{x})
= \mathcal{N}\bigl(\boldsymbol{\mu}_{\phi}(\mathbf{x}),
\operatorname{diag}(\boldsymbol{\sigma}_{\phi}^{2}(\mathbf{x}))\bigr),
\quad
\mathbf{z}=\boldsymbol{\mu}_{\phi}(\mathbf{x})
+\boldsymbol{\sigma}_{\phi}(\mathbf{x})\odot \boldsymbol{\epsilon},
\quad
\boldsymbol{\epsilon}\sim\mathcal{N}(\mathbf{0},\mathbf{I})
\end{equation}
The decoder $\mathrm{Dec}_{\theta}$ reconstructs the video as
$\hat{\mathbf{x}}=\mathrm{Dec}_{\theta}(\mathbf{z})$. The autoencoder is optimized with
pixel reconstruction, perceptual, adversarial, and KL regularization losses:
\begin{equation}
\mathcal{L}_{\mathrm{VAE}}
=
\lambda_{\mathrm{rec}}\|\hat{\mathbf{x}}-\mathbf{x}\|_{1}
+\lambda_{\mathrm{lpips}}\mathcal{L}_{\mathrm{LPIPS}}(\hat{\mathbf{x}},\mathbf{x})
+\lambda_{\mathrm{gan}}\mathcal{L}_{\mathrm{GAN}}(\hat{\mathbf{x}},\mathbf{x})
+\beta\,
D_{\mathrm{KL}}\bigl(q_{\phi}(\mathbf{z}\mid\mathbf{x})
\;\|\; \mathcal{N}(\mathbf{0},\mathbf{I})\bigr)
\end{equation}
where the first three terms encourage faithful and perceptually realistic video
reconstruction, and the KL term regularizes the latent distribution. This
standard objective defines a compact reconstructive latent space, but does not
explicitly constrain what geometric structure the latent should preserve.

\noindent\textbf{Video Geometry Formation.}
A video can be viewed as the 2D projection of a dynamic 3D scene. This
formation process suggests three natural levels of geometric grounding. First,
each frame is tied to an underlying 3D layout, which we represent by a depth map
$D_i$ for frame $I_i$. Second, as the scene and camera evolve, visible points
induce dense correspondence across frames, represented by optical flow
$F_{i\rightarrow j}$, where a pixel $\mathbf{p}$ in frame $i$ is matched to
$\mathbf{p}'=\mathbf{p}+F_{i\rightarrow j}(\mathbf{p})$ in frame $j$.
Third, for static scene points, depth and camera poses define a rigid projective
mapping between views. This mapping accounts for camera motion, but not independent
object motion. To reconstruct a target frame $I_j$ from a source frame $I_i$, we
use target-driven backward warping. For simplicity, let $\mathbf{K}$ denote shared
camera intrinsics, and let
$\mathbf{G}_{j\rightarrow i}=[\mathbf{R}_{j\rightarrow i}|\mathbf{t}_{j\rightarrow i}]$
transform points from the target camera coordinates to the source camera coordinates.
A target pixel $\mathbf{p}_j$ with homogeneous coordinate $\tilde{\mathbf{p}}_j$
and depth $D_j(\mathbf{p}_j)$ is back-projected as
\begin{equation}
\mathbf{X}_j(\mathbf{p}_j)
=D_j(\mathbf{p}_j)\mathbf{K}^{-1}\tilde{\mathbf{p}}_j
\end{equation}
The corresponding point in the source camera coordinates and its projected
sampling location are
\begin{equation}
\label{eq:projective_mapping}
\mathbf{X}_i
=\mathbf{R}_{j\rightarrow i}\mathbf{X}_j(\mathbf{p}_j)
+\mathbf{t}_{j\rightarrow i},
\qquad
\mathbf{p}_i=\Pi(\mathbf{K}\mathbf{X}_i)
\end{equation}
where $\Pi([x,y,z]^{\top})=(x/z,y/z)^{\top}$. This defines the target-to-source coordinate map:
\begin{equation}
\label{eq:view_warp}
\mathbf{p}_i
=
\mathcal{W}_{\mathrm{view}}
(\mathbf{p}_j,D_j(\mathbf{p}_j),\mathbf{G}_{j\rightarrow i},\mathbf{K})
\end{equation}
Backward warping samples the source at $\mathbf{p}_i$ to populate the target
location $\mathbf{p}_j$, using target-to-source coordinates for source-to-target reconstruction.
\subsection{Video Geometry Decoder}
\label{sec:video_geometry_decoder}

Videos are observations of dynamic 3D worlds, and their latent
representations should therefore capture the geometric structure
underlying video formation. Motivated by this perspective, VGGA
introduces a \emph{Video Geometry Decoder} that explicitly incorporates
geometric grounding into the decoding process, rather than treating
the decoder solely as an RGB reconstruction module.

A key challenge arises from the spatio-temporal compression performed
by video encoders. Given an input video, the resulting latent
representation
$\mathbf{z}\in\mathbb{R}^{(1+T/r_t)\times H/r_s\times W/r_s\times C}$
is compressed along both spatial and temporal dimensions. In particular,
when $r_t>1$, the latent sequence is not directly aligned with
frame-level geometric signals such as depth,
and projective geometry.

To bridge this mismatch, VGGA employs a shared spatio-temporal
upsampling trunk $U_{\theta}$ that progressively decodes
$\mathbf{z}$ into a frame-aligned intermediate representation:
\begin{equation}
\mathbf{h}
=
U_{\theta}(\mathbf{z})
=
U^{\mathrm{st}}_{2}
\bigl(U^{\mathrm{st}}_{1}(\mathbf{z})\bigr),
\end{equation}
where $U^{\mathrm{st}}$ denotes a spatio-temporal upsampling block
that increases both temporal and spatial resolution. The resulting
feature representation
$\mathbf{h}=\{\mathbf{h}_t\}_{t=0}^{T_h}
\in\mathbb{R}^{(1+T_h)\times H_h\times W_h\times C_h}$
has a higher spatio-temporal resolution than $\mathbf{z}$. In our
implementation, we set $T_h=T$ to establish frame-level temporal
alignment with the input video. The shared representation $\mathbf{h}$ is subsequently used for both
RGB reconstruction and geometric grounding. RGB reconstruction is
performed by an RGB head,
$\hat{\mathbf{x}}=R_{\theta}(\mathbf{h})$.

\subsection{Video Geometry Grounding}
\label{sec:geometry_grounding}
To ground video latents in the geometry of dynamic 3D worlds, VGGA introduces
three complementary grounding tasks. \emph{Structural Depth Grounding} guides the latent to encode
per-frame 3D layout. \emph{Motion
Correspondence Grounding} preserves dense cross-frame correspondence. \emph{Projective Warping Grounding} further
regularizes the latent with camera-induced view transformation, requiring
decoded features to be consistent under projective geometry. 

\noindent\textbf{Structural Depth Grounding.}
Depth provides the most direct cue for the per-frame 3D layout of a video. We
therefore use it as the first grounding signal to make the latent
$\mathbf{z}$ aware of scene structure through the shared decoded feature
$\mathbf{h}$. 
Specifically, our depth grounding requires each frame-level feature $\mathbf{h}_t$ to predict a
depth map:
\begin{equation}
\hat{\mathbf{D}}_t=H_d(\mathbf{h}_t)
\end{equation}
Let $\mathbf{D}$ be the target depth map. Following Video Depth
Anything~\citep{chen2025video}, we use a standard video depth loss $\mathcal{L}_{\mathrm{depth}}$ for supervision (see~\Cref{apd:loss} for details).

\noindent\textbf{Motion Correspondence Grounding.}
Structural depth grounds the layout of individual frames, but it does not
specify how scene points correspond over time. To make the latent aware of video
dynamics, we further introduce \textit{Motion Correspondence Grounding} based on dense
optical flow. This task requires neighboring frame-level features to predict the
displacement of visible content, encouraging $\mathbf{z}$ to retain the
cross-frame correspondence needed for coherent motion. Given neighboring decoded
features, the flow head predicts forward and backward flow:
\begin{equation}
\hat{\mathbf{F}}_{t\rightarrow t+1},\hat{\mathbf{F}}_{t+1\rightarrow t}
=H_f(\mathbf{h}_t,\mathbf{h}_{t+1})
\end{equation}
Let $\mathbf{F}$ be the target flow. Following
RAFT~\citep{teed2020raft}, we use a simple $l_1$ endpoint regression loss for flow
grounding $\mathcal{L}_{\mathrm{flow}}$ (see~\Cref{apd:loss} for details).
\begin{figure}[!t]
  \centering
  \includegraphics[width=\linewidth]{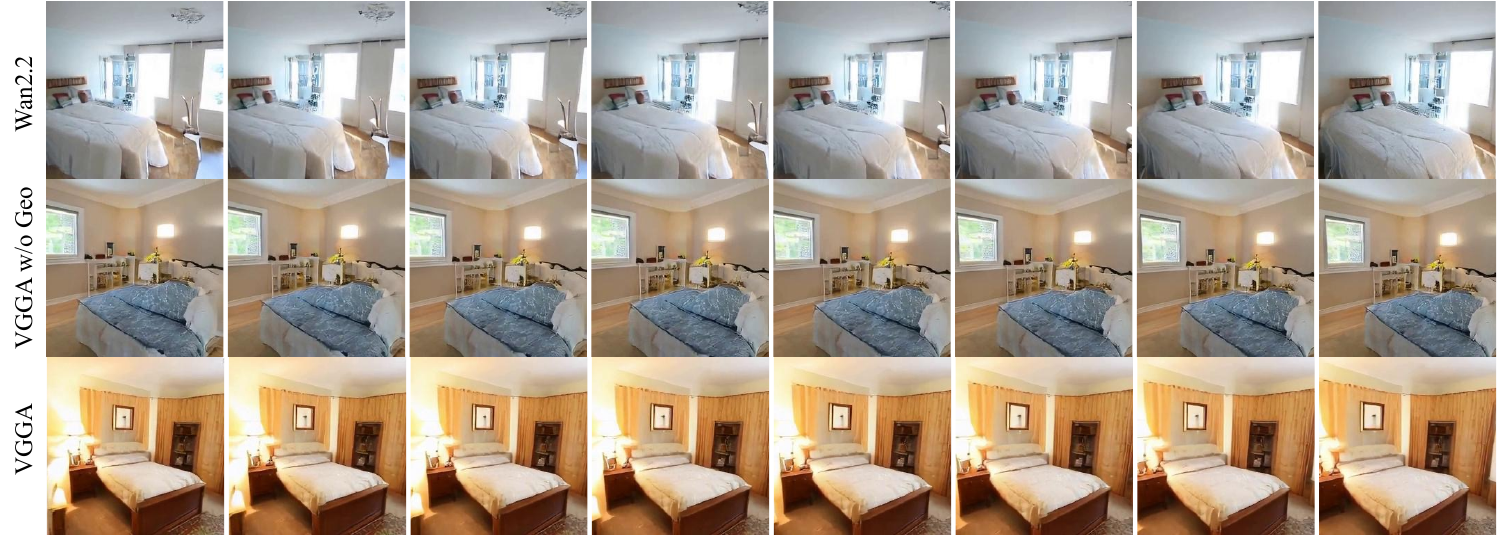}
  \vspace{-0.8em}
  \caption{\textbf{Qualitative comparison on RealEstate10K.}
  Compared to Wan2.2 VAE and our reconstruction-only baseline (w/o Geo), 
  VGGA synthesizes videos with more physically plausible view changes and robust structural stability. All models are evaluated under the same unconditional setting.}
  \label{fig:re10k_vis}
  \vspace{-0.8em}
\end{figure}

\noindent\textbf{Projective Warping Grounding.}  
We further enforce consistency between feature-space and RGB-space warping:  
features warped under a given camera transformation should decode to the RGB frame  
warped by the same geometry.  
This objective encourages the shared reconstruction features  
to preserve camera-aware geometric equivariance.

For a source frame $I_i$ and target frame $I_j$,  
we construct a backward sampling map in \Cref{eq:view_warp}  
using the target-frame VGGT pseudo depth $\mathbf{D}_j$  
and the target-to-source camera transformation  
$\mathbf{G}_{j\rightarrow i}$.
The same view-warping operation is applied at both feature  
and RGB resolutions.  
Let $\mathbf{D}_j^h$ and $\mathbf{K}_h$ denote the pseudo depth  
and camera intrinsics at the shared feature resolution, respectively.  
The target-aligned feature and its decoded RGB reconstruction are
\begin{equation}
\tilde{\mathbf{h}}_{i\rightarrow j}
=
\operatorname{Sample}\!\left(
\mathbf{h}_i,
\mathcal{W}_{\mathrm{view}}
(\cdot,\mathbf{D}_j^h,\mathbf{G}_{j\rightarrow i},\mathbf{K}_h)
\right),
\qquad
\hat{I}^{\mathrm{warp}}_{i\rightarrow j}
=
R_{\theta}(\tilde{\mathbf{h}}_{i\rightarrow j}).
\end{equation}

Correspondingly, the RGB supervision target is obtained  
by applying the same geometric transformation at image resolution:
\begin{equation}
I^{\mathrm{ref}}_{i\rightarrow j}
=
\operatorname{Sample}\!\left(
I_i,
\mathcal{W}_{\mathrm{view}}
(\cdot,\mathbf{D}_j,\mathbf{G}_{j\rightarrow i},\mathbf{K})
\right),
\end{equation}
where $\mathbf{D}_j$ and $\mathbf{K}$ are defined at RGB resolution. 
Following reconstruction,  we use a simple $l_1$ loss for projective warping grounding $\mathcal{L}_{\mathrm{warp}}$ (see~\Cref{apd:loss} for details).

\subsection{Hybrid Geometry Supervision on Unlabelled Videos}
\label{sec:scalable_grounding}

Scaling geometry-grounded video autoencoding is challenging,  
as large-scale video datasets rarely provide explicit geometric annotations.  
Dense depth and motion labels are costly to acquire, while reliable camera poses  
are particularly difficult to obtain for dynamic videos.  
VGGA therefore adopts a hybrid supervision strategy that exploits geometric cues  
according to their availability.

For general videos, we use pseudo depth and optical flow predicted by  
off-the-shelf models to supervise \emph{Structural Depth Grounding} and  
\emph{Motion Correspondence Grounding} 
; sensitivity to pseudo-label quality is analyzed in
\Cref{apd:sensitivity}.

In contrast, \emph{Projective Warping Grounding} requires reliable camera poses  
and assumes static scene geometry. We therefore apply this objective only to  
a small subset of static-scene videos with known camera poses, where projective  
view transformations can be reliably established.
The three grounding objectives are jointly optimized with standard VAE reconstruction objective:
\begin{equation}
\mathcal{L}_{\mathrm{VGGA}}
=
\mathcal{L}_{\mathrm{VAE}}
+
\lambda_d \mathcal{L}_{\mathrm{depth}}
+
\lambda_f \mathcal{L}_{\mathrm{flow}}
+
\lambda_w \mathcal{L}_{\mathrm{warp}},
\end{equation}
where $\mathcal{L}_{\mathrm{warp}}$ is applied only when projective supervision  
is available.

Through the shared Video Geometry Decoder, camera-aware priors learned from  
static scenes are jointly integrated with depth and motion cues from general videos.  
This hybrid training enables projective geometric priors to transfer beyond  
the statically supervised data and benefit dynamic video representation  
and generation (see \Cref{fig:geometry_analysis}(b)).

\section{Experiments}
\subsection{Setup}

\noindent\textbf{Architecture.}
The video VAE backbone of VGGA adopts the same architectural design as Wan2.2 VAE~\citep{wan2025wan}.
To isolate the effect of geometry grounding, both VGGA and its reconstruction-only baseline, VGGA w/o Geo, are trained from scratch on identical datasets.
The Video Geometry Decoder
adds lightweight geometry branches on top of the shared spatio-temporal
upsampling trunk. The depth head consists of two spatial upsampling blocks,
while the flow head follows the optical-flow prediction design of
FlowNet2~\citep{ilg2017flownet}.

\begin{table}[t]
\footnotesize
\centering
\vspace{-2mm}
\caption{\textbf{Comparison of generation performance at 17-frame $256\times256$ resolution.} The best and second-best
results are indicated in \textbf{bold} and \underline{underlined}. The notation
$tT sS cC$ denotes a temporal downsampling factor of $T$, a spatial
downsampling factor of $S\times S$, and a latent channel dimension of $C$. ``VGGA w/o Geo'' denotes the reconstruction-only baseline trained without geometry grounding.}
\vspace{-3mm}
\label{tab:ucf_realestate_generation}
\setlength{\tabcolsep}{3.5pt}
\resizebox{\linewidth}{!}{%
\begin{tabular}{lccccccccc}
\toprule
\multirow{2}{*}{\textbf{Method}} & \multirow{2}{*}{\textbf{Latent config}} &
\multicolumn{3}{c}{\textbf{RealEstate10K}} &
\multicolumn{3}{c}{\textbf{UCF101}} &
\multicolumn{2}{c}{\textbf{Taichi-HD}} \\
\cmidrule(lr){3-5}\cmidrule(lr){6-8}\cmidrule(lr){9-10}
& & \textbf{FVD$\downarrow$} & \textbf{KVD$\downarrow$} & \textbf{Met3R$\downarrow$}
& \textbf{FVD$\downarrow$} & \textbf{KVD$\downarrow$} & \textbf{IS$\uparrow$}
& \textbf{FVD$\downarrow$} & \textbf{KVD$\downarrow$} \\
\midrule
\textbf{CogX-VAE}~\citep{yang2024cogvideox} & t4s8c16 & 94.12 & 10.41 & - & 176.90 & 16.47 & 64.19 & - & - \\
\textbf{IV-VAE}~\citep{wu2025improved} & t4s8c16 & 92.37 & 8.35 & - & 175.74 & 22.32 & 64.51 & - & - \\
\textbf{WF-VAE}~\citep{li2025wf} & t4s8c16 & 107.26 & 12.56 & - & 188.19 & 33.01 & 67.49 & - & - \\
\textbf{Hunyuan-VAE}~\citep{kong2024hunyuanvideo} & t4s8c16 & \underline{83.45} & 13.23 & - & 210.30 & 52.81 & 66.40 & - & - \\
\textbf{Wan2.1 VAE}~\citep{wan2025wan} & t4s8c16 & 83.84 & 10.64 & - & 167.10 & \textbf{11.54} & 66.04 & - & - \\
\rowcolor{blue!8}
\textbf{VGGA w/o Geo} & t4s8c16 & 88.21 & \underline{8.06} & - & \underline{163.21} & 21.35 & \underline{67.36} & - & - \\
\rowcolor{blue!8}
\textbf{VGGA} & t4s8c16 & \textbf{69.17} & \textbf{5.02} & - & \textbf{148.15} & \underline{16.21} & \textbf{68.75} & - & - \\
\midrule
\textbf{Wan2.2 VAE}~\citep{wan2025wan} & t4s16c48 & 87.15 & 10.11 & 0.0494 & 180.79 & 17.80 & 67.32 & 218.92 & 109.58 \\
\textbf{SSVAE}~\citep{liu2025delving} & t4s16c48 & \underline{79.08} & 8.79 & \underline{0.0487} & 168.68 & 19.71 & 66.39 & \underline{189.26} & \underline{79.37} \\
\rowcolor{blue!8}
\textbf{VGGA w/o Geo} & t4s16c48 & 89.49 & \underline{4.03} & 0.0514 & \underline{163.78} & \underline{18.72} & \underline{68.14} & 195.44 & 80.34 \\
\rowcolor{blue!8}
\textbf{VGGA} & t4s16c48 & \textbf{64.31} & \textbf{3.32} & \textbf{0.0421} & \textbf{141.27} & \textbf{16.05} & \textbf{70.32} & \textbf{121.76} & \textbf{46.89} \\
\bottomrule
\end{tabular}%
}
\end{table}

\begin{table}[tp]
\centering
\footnotesize
\vspace{-3mm}
\caption{\textbf{Reconstruction performance on the Kinetics-400 validation set at different resolutions.}  
Unlike SSVAE, which substantially degrades reconstruction quality relative to Wan2.2 VAE,  
VGGA preserves reconstruction fidelity close to its reconstruction-only baseline  
(``VGGA w/o Geo''), demonstrating the strong compatibility between geometric grounding  
and VAE reconstruction. Positive (green) values denote improvements, while negative (red)  
values denote degradations.}
\vspace{-3mm}
\label{tab:kinetics400_reconstruction}
\setlength{\tabcolsep}{4.0pt}
\resizebox{\linewidth}{!}{%
\begin{tabular}{lcccccccc}
\toprule
\multirow{2}{*}{\textbf{Method}} &
\multicolumn{4}{c}{$17\times256\times256$} &
\multicolumn{4}{c}{$17\times512\times512$} \\
\cmidrule(lr){2-5}\cmidrule(lr){6-9}
& \textbf{rFVD$\downarrow$} & \textbf{PSNR$\uparrow$} & \textbf{SSIM$\uparrow$} & \textbf{LPIPS$\downarrow$}
& \textbf{rFVD$\downarrow$} & \textbf{PSNR$\uparrow$} & \textbf{SSIM$\uparrow$} & \textbf{LPIPS$\downarrow$} \\
\midrule
\textbf{CogX-VAE}~\citep{yang2024cogvideox} & 4.90 & 33.78 & 0.97 & 0.027 & 1.79 & 36.00 & 0.99 & 0.024 \\
\textbf{IV-VAE}~\citep{wu2025improved} & 2.78 & 34.08 & 0.96 & 0.019 & 0.97 & 37.24 & 0.96 & 0.016 \\
\textbf{WF-VAE-L}~\citep{li2025wf} & 3.06 & 33.48 & 0.96 & 0.023 & 1.08 & 35.93 & 0.96 & 0.023 \\
\textbf{Hunyuan-VAE}~\citep{kong2024hunyuanvideo} & 2.96 & 34.30 & 0.97 & 0.016 & 0.90 & 37.13 & 0.97 & 0.015 \\
\textbf{Wan2.1 VAE}~\citep{wan2025wan} & 2.92 & 33.21 & 0.95 & 0.018 & 1.02 & 36.15 & 0.97 & 0.017 \\
\midrule
\textbf{Wan2.2 VAE}~\citep{wan2025wan} & 3.42 & 33.78 & 0.96 & 0.015 & 1.22 & 36.75 & 0.97 & 0.015 \\
\textbf{SSVAE}~\citep{liu2025delving} & 7.50 & 31.18 & 0.96 & 0.036 & 2.16 & 34.45 & 0.97 & 0.028 \\
Gain vs. Wan2.2 VAE & \textcolor{red}{$-4.08$} & \textcolor{red}{$-2.60$} & 0.00 & \textcolor{red}{$-0.021$} & \textcolor{red}{$-0.94$} & \textcolor{red}{$-2.30$} & 0.00 & \textcolor{red}{$-0.013$} \\
\midrule
\rowcolor{blue!8}
\textbf{VGGA w/o Geo} & 3.56 & 33.61 & 0.97 & 0.019 & 1.34 & 36.65 & 0.97 & 0.017 \\
\rowcolor{blue!8}
\textbf{VGGA} & 3.42 & 33.52 & 0.97 & 0.018 & 1.30 & 36.47 & 0.97 & 0.019 \\
\rowcolor{blue!8}
Gain vs. VGGA w/o Geo & \textcolor{green!50!black}{$+0.14$} & \textcolor{red}{$-0.09$} & 0.00 & \textcolor{green!50!black}{$+0.001$} & \textcolor{green!50!black}{$+0.04$} & \textcolor{red}{$-0.18$} & 0.00 & \textcolor{red}{$-0.002$} \\
\bottomrule
\end{tabular}%
}
\vspace{-4mm}
\end{table}

\noindent\textbf{Training Data.}
We train VGGA on a mixture of a curated high-quality 10M-video dataset and
RealEstate10K~\citep{zhou2018stereo}, sampled at a 4:1 ratio. The 10M videos do
not require manual depth or flow annotations. We obtain depth pseudo labels
using VGGT~\citep{wang2025vggt} and optical-flow pseudo labels using RAFT~\citep{
teed2020raft}. Structural Depth Grounding and Motion Correspondence Grounding
are applied to all training videos. Projective Warping Grounding is applied only
to RealEstate10K samples. Additional training details are provided in~\Cref{apd:imp}.

\noindent\textbf{Evaluation.}
We evaluate generation on RealEstate10K~\citep{zhou2018stereo}, UCF101~\citep{soomro2012ucf101}, and
Taichi-HD~\citep{siarohin2019first}. RealEstate10K contains strong geometric
structure and camera motion, and is used to assess view consistency and
camera-motion stability. UCF101 evaluates
action-conditioned generation, while Taichi-HD evaluates unconditional human
motion generation. For reconstruction, we evaluate it on the
Kinetics-400 validation set~\citep{carreira2017quo}. 
Additional evaluation details are provided in~\Cref{apd:imp}.

\subsection{Video Generation and Reconstruction}

\noindent\textbf{Geometry-centric generation.}  
We evaluate VGGA on RealEstate10K to assess its ability to model complex camera motion  
and projective view changes. As shown in \Cref{tab:ucf_realestate_generation},  
VGGA achieves the best FVD of 64.31, outperforming SSVAE by 14.77 points under the  
same $t4s16c48$ latent configuration and reducing FVD by 25.18 points over the  
reconstruction-only baseline. VGGA also improves geometric consistency, reducing  
Met3R from 0.0514 to 0.0421. Together with the qualitative results in  
\Cref{fig:re10k_vis}, these results demonstrate that geometry grounding improves  
both generation quality and geometric fidelity under complex view transformations.

\noindent\textbf{Action-conditioned generation.}
Beyond camera-centric scenarios, we evaluate whether geometry-grounded latents support diverse human actions. 
As reported in \Cref{tab:ucf_realestate_generation}, VGGA achieves an FVD of 141.27 and an Inception Score (IS) 
of 70.32 on UCF101. Under the t4s16c48 configuration, VGGA reduces FVD by 27.41 points relative to SSVAE (168.68) 
and by 22.51 points relative to its reconstruction-only counterpart (163.78). Under t4s8c16, VGGA likewise improves 
FVD from 163.21 to 148.15. These results demonstrate that geometry grounding benefits action-conditioned 
synthesis across both latent configurations.

\noindent\textbf{Human-motion generation.}
Furthermore, we extend our evaluation to Taichi-HD to test the model's capability in handling articulated 
human motion, which demands that the latent distribution encapsulate both fine-grained local deformations 
and global temporal coherence. As shown in \Cref{tab:ucf_realestate_generation}, VGGA reduces 
FVD from 218.92 (Wan2.2 VAE) to 121.76 under the same t4s16c48 configuration. Compared to its reconstruction-only 
counterpart, VGGA reduces FVD by 73.68 points (195.44 to 121.76). These results substantiate the effectiveness of 
VGGA across different video generation tasks.

\begin{figure}[!t]
  \centering
  \includegraphics[width=\linewidth]{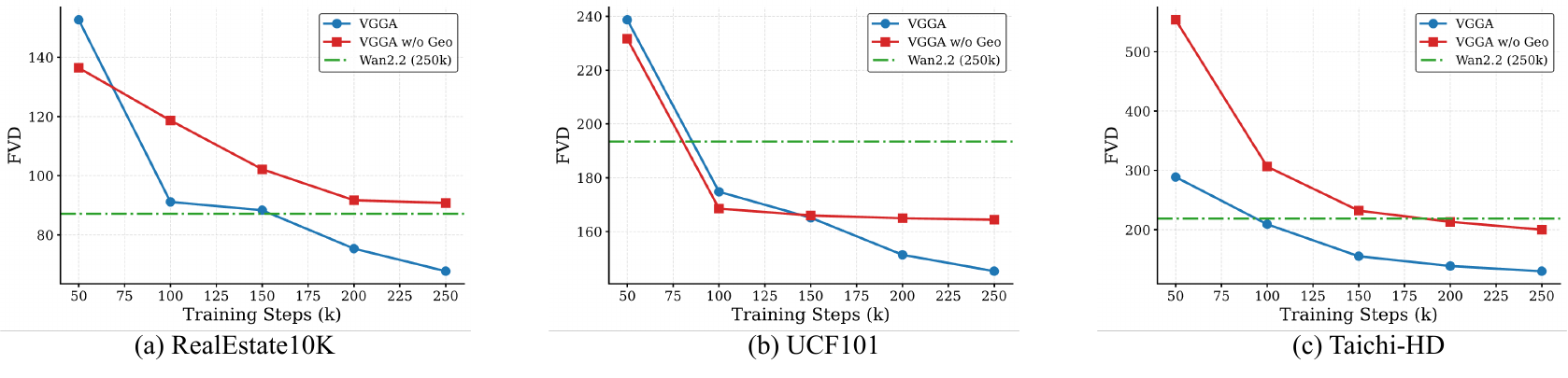}
  \vspace{-6mm}
  \caption{\textbf{Effect of geometry grounding on Video DiT training.}
  We compare VGGA with its reconstruction-only counterpart (VGGA w/o Geo) on
  three generation benchmarks. Geometry grounding consistently yields lower FVD throughout
  training, indicating more favorable optimization dynamics and improved
  diffusability of the resulting video latents.}
  \label{fig:fvd_steps}
  \vspace{-3mm}
\end{figure}

\begin{figure}[tp]
\centering
\captionsetup{skip=2pt}
\includegraphics[width=\linewidth]{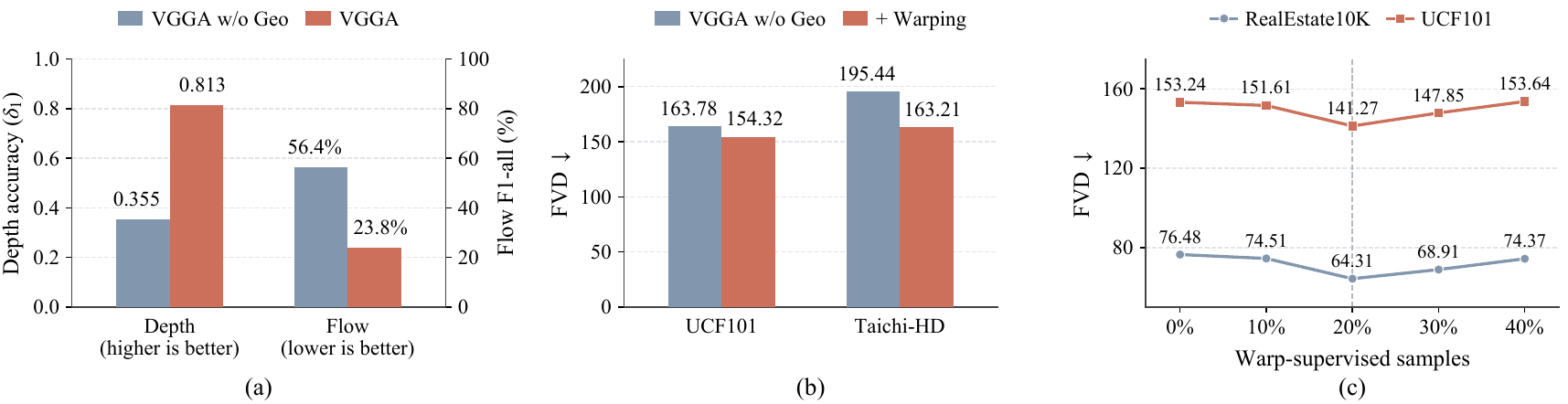}
\caption{\textbf{Geometric representation learning and transfer from static-scene supervision.}  
\textbf{(a) Latent geometric probing:} linear probes on frozen latents recover depth  
and optical flow on the RealEstate10K test split, evaluated against VGGT and RAFT  
pseudo-labels, respectively.  
\textbf{(b) Transfer to dynamic generation:} adding only \emph{Projective Warping Grounding}  
on static-scene RealEstate10K (``+ Warping'') improves generation on dynamic-scene  
datasets, including UCF101 and Taichi-HD.  
\textbf{(c) Warp-supervision proportion:} a 20\% proportion of warp-supervised samples  
(our default, dashed line) achieves the lowest FVD on both datasets.}
\label{fig:geometry_analysis}
\vspace{-5mm}
\end{figure}

\noindent\textbf{Video reconstruction.}  
As shown in \Cref{tab:kinetics400_reconstruction}, VGGA maintains reconstruction  
quality close to its reconstruction-only baseline across resolutions.  
At $17\times256\times256$, rFVD improves from 3.56 to 3.42 and LPIPS from  
0.019 to 0.018, with only a 0.09 dB drop in PSNR. At $17\times512\times512$,  
rFVD improves from 1.34 to 1.30, while PSNR decreases by 0.18 dB and LPIPS  
increases slightly from 0.017 to 0.019. SSIM remains unchanged at 0.97.  
In contrast, SSVAE incurs PSNR drops of 2.60 and 2.30 dB relative to Wan2.2 VAE.  
These results show that VGGA substantially preserves reconstruction fidelity,  
demonstrating the compatibility of geometric grounding with VAE reconstruction.

\subsection{Analysis}
We analyze how geometry grounding affects latent diffusability, the recoverability
of geometric information, and the transfer of static-scene supervision to dynamic
video generation.

\noindent\textbf{Diffusability of video latents.}
We evaluate VAE's diffusability by tracking FVD during the training of identical Latte-XL/1 generators. 
As shown in \Cref{fig:fvd_steps}, VGGA consistently achieves lower FVD and faster convergence 
than the reconstruction-only baseline. This suggests that geometry grounding regularizes the latent 
space with explicit structural cues, thereby reducing the complexity of the downstream generative task 
and rendering the latents more diffusion-friendly.

\noindent\textbf{Latent geometric probing.}
We train linear depth and optical-flow probes on frozen video latents  
and evaluate them on the RealEstate10K test split, using VGGT and RAFT  
pseudo-labels as reference targets. As shown in  
\Cref{fig:geometry_analysis}(a), VGGA improves depth $\delta_1$ from  
0.355 to 0.813 and reduces the flow outlier rate (F1-all) from  
56.4\% to 23.8\%. These results demonstrate that VGGA encodes  
substantially stronger geometric structure and motion correspondence,  
making such information more linearly accessible from the latent space.

\noindent\textbf{Transfer from static scenes to dynamic generation.}  
To isolate the contribution of static-scene supervision, we add only the  
projective warping objective on RealEstate10K to the reconstruction-only baseline.  
As shown in \Cref{fig:geometry_analysis}(b), FVD decreases from 163.78 to 154.32  
on UCF101 and from 195.44 to 163.21 on Taichi-HD.
These results demonstrate that camera-aware geometric priors learned from static scenes  
transfer effectively to dynamic video generation and provide complementary benefits  
to depth and motion supervision.

\begin{table}[t]
\centering
\begin{minipage}[t]{0.48\linewidth}
\centering
\caption{{Ablation on geometry grounding.}}
\vspace{-3mm}
\label{tab:ablation}
\scriptsize
\setlength{\tabcolsep}{1pt}
\resizebox{\linewidth}{!}{
\begin{tabular}{lcccc}
\toprule
\multirow{2}{*}{\textbf{Components}} & \multicolumn{2}{c}{\textbf{Generation}} &
\multicolumn{2}{c}{\textbf{Reconstruction}} \\
\cmidrule(lr){2-3}\cmidrule(lr){4-5}
& \textbf{FVD$\downarrow$} & \textbf{KVD$\downarrow$}
& \textbf{rFVD$\downarrow$} & \textbf{PSNR$\uparrow$} \\
\midrule
\textbf{VGGA w/o Geo} & 163.78 & 18.72 & 3.56 & 33.61 \\
w/o Depth Grounding & 156.31 & 18.10 & \textbf{3.38} & 33.50\\
w/o Motion Grounding & 146.43 & 16.96 & 3.50 & \textbf{33.63} \\
w/o Warping Grounding & 152.24 & 17.78 & 3.58 & 33.58 \\
\textbf{VGGA full} & \textbf{141.27} & \textbf{16.05} & 3.42 & 33.52 \\
\bottomrule
\end{tabular}
}
\end{minipage}
\hfill
\begin{minipage}[t]{0.48\linewidth}
\centering
\caption{{Ablation on geometry branch position.}}
\vspace{-3mm}
\label{tab:sharing_position}
\scriptsize
\setlength{\tabcolsep}{0.5pt}
\resizebox{\linewidth}{!}{
\begin{tabular}{lccccc}
\toprule
\multirow{2}{*}{\textbf{Components}} &
\multirow{2}{*}{\shortstack[c]{\textbf{Frame-}\\\textbf{aligned}}} &
\multicolumn{2}{c}{\textbf{Generation}} &
\multicolumn{2}{c}{\textbf{Reconstruction}} \\
\cmidrule(lr){3-4}\cmidrule(lr){5-6}
& & \textbf{FVD$\downarrow$} & \textbf{KVD$\downarrow$}
& \textbf{rFVD$\downarrow$} & \textbf{PSNR$\uparrow$} \\
\midrule
Separate branch & \ding{55}
& 159.12 & 18.80 & 3.56 & \textbf{33.60} \\
after $U^{\mathrm{st}}_{1}$ & \ding{55}
& 149.27 & 18.15 & 3.48 & 33.31 \\
after $U^{\mathrm{st}}_{2}$ (VGGA) & \ding{51}
& \textbf{141.27} & \textbf{16.05} & \textbf{3.42} & 33.52 \\
after $U^{\mathrm{s}}_{3}$ & \ding{51}
& 153.46 & 18.66 & 4.31 & 32.14 \\
after $U^{\mathrm{s}}_{4}$ & \ding{51}
& - & - & - & 12.08 \\
\bottomrule
\end{tabular}
}
\end{minipage}
\vspace{-5mm}
\end{table}

\subsection{Ablation Study}

\noindent\textbf{Geometry grounding signals.}  
The three grounding signals provide complementary constraints on scene structure,  
temporal correspondence, and camera-induced geometry. As shown in  
\Cref{tab:ablation}, removing geometric supervision increases FVD from 141.27  
to 163.78, while all leave-one-out variants degrade generation performance.  
This confirms that the three geometric cues contribute complementarily to  
learning generation-friendly video latents.

\noindent\textbf{Shared branching position.}  
As shown in \Cref{tab:sharing_position}, branching after  
$U^{\mathrm{st}}_2$ achieves the best overall performance. Earlier branching  
lacks sufficient frame-level alignment, whereas later branching operates on  
increasingly RGB-specific features. A separate geometry branch is also less  
effective, confirming that geometric supervision is most effective when applied  
to the shared decoded representation.

\noindent\textbf{Proportion of warp-supervised samples.}  
\Cref{fig:geometry_analysis}(c) studies the proportion of samples receiving  
projective warping supervision. Increasing this proportion from 0\% to 20\%  
reduces FVD from 76.48 to 64.31 on RealEstate10K and from 153.24 to 141.27  
on UCF101. Further increasing the proportion to 30\% or 40\% degrades  
performance on both datasets. Thus, the default 20\% mixture provides the  
best trade-off, introducing effective camera-aware supervision while retaining  
general videos as the majority of training data.

\section{Conclusion}

We presented \textbf{VGGA},  
which introduces a geometry-aware perspective for learning generation-friendly  
video latents by grounding autoencoding in the geometric structure of dynamic 3D scenes.  
VGGA enables geometric learning to be effectively integrated with reconstruction  
and scaled to large-scale unlabelled videos.  
Extensive experiments demonstrate that geometry-grounded autoencoding  
substantially improves downstream video generation while preserving reconstruction quality.

\noindent\textbf{Limitations.} VGGA focuses on frame-level and short-range geometric consistency,  
without explicitly modeling long-horizon dynamics or full 3D scene structure.  
Extending geometry grounding to longer temporal ranges and scene-level 3D modeling  
is a promising direction for future work.
\newpage

\section*{Use of Large Language Models (LLMs)}
\label{sup:LLM}
In preparing this manuscript, we used large language models solely as a lightweight writing aid for grammar, 
wording, and formatting suggestions. The models were \emph{not} used to generate research ideas, design algorithms, 
write code, run experiments, analyze data, or draft scientific content. All technical claims, methods, 
and conclusions were conceived, produced, and verified by the authors. Suggested edits from LLMs were manually 
reviewed and integrated at the authors' discretion. 
And we accept full responsibility for the accuracy and integrity of the manuscript, including ensuring that 
no plagiarized or misrepresented content from a LLM is included.

\bibliographystyle{iclr2027_conference}
\bibliography{iclr2027_conference}

\newpage

\begin{center}
    {\LARGE \textbf{Appendix}}
\end{center}

\appendix

\section{Implementation Details}
\label{apd:imp}
This section provides additional implementation details omitted from the main
paper for clarity. We describe the evaluation metrics, pseudo-label generation
for geometry grounding, autoencoder training losses, and downstream latent
generator training protocol.

\subsection{Evaluation Metrics}
For generation, we follow common video-generation evaluation protocols and
compute Fr\'echet Video Distance (FVD) and Kernel Video Distance
(KVD)~\citep{unterthiner2018towards} in the I3D feature space. FVD measures the
Fr\'echet distance between Gaussian-fitted real and generated video features,
while KVD computes a kernel two-sample distance between the two feature sets.
We additionally report Inception Score (IS)~\citep{salimans2016improved} on
datasets where the metric is applicable. Unless otherwise specified, generation
metrics are computed with 2048 generated videos at
$17\times256\times256$ resolution.

For geometric consistency, we follow Met3R~\citep{asim2025met3r} and evaluate
3D consistency between adjacent frames of generated videos on RealEstate10K
to obtain video-level Met3R scores. Lower scores indicate better geometric consistency.

We evaluate reconstruction using rFVD, peak signal-to-noise ratio
(PSNR)~\citep{huynh2008scope}, structural similarity
(SSIM)~\citep{wang2004image}, and learned perceptual image patch similarity
(LPIPS)~\citep{zhang2018unreasonable}. Reconstruction FVD (rFVD) is computed between original and
reconstructed video clips using the same feature extractor as FVD. PSNR, SSIM,
and LPIPS are computed frame-wise and averaged over all frames and validation
videos.

\subsection{Pseudo Geometric Supervision}
VGGA does not require manual depth or optical-flow annotations. For the curated
10M-video dataset and RealEstate10K, we generate pseudo depth maps using
VGGT~\citep{wang2025vggt} and pseudo optical flow using RAFT~\citep{
teed2020raft}. Depth and flow pseudo labels are produced offline and resized to
the training resolution. We mix the curated 10M-video dataset and RealEstate10K
with a 4:1 sampling ratio. For projective warping, we use RealEstate10K samples
with valid camera poses and predominantly static content. Target-frame VGGT
pseudo depth and target-to-source camera transformations define the backward
sampling maps; depth maps and intrinsics are resized and scaled to the decoded
feature grid, respectively. During mixed-data training,
Structural Depth Grounding and Motion Correspondence Grounding are applied to
all samples, while Projective Warping Grounding is applied only to RealEstate10K
samples with valid camera poses.
Sensitivity to alternative depth and flow teachers is evaluated in
\Cref{apd:teacher_robustness}.

\subsection{Autoencoder Training Losses}
\label{apd:loss}

\noindent\textbf{Structural Depth Grounding Loss.}
For depth regression, we transform the pseudo depth to
$\mathbf{D}'=\log(1+\mathbf{D})$ element-wise and denote the corresponding
prediction in this domain by $\hat{\mathbf{D}}'$. Following Video Depth
Anything~\citep{chen2025video}, we use a video depth loss $\mathcal{L}_{\mathrm{depth}}$ that combines
depth regression, spatial gradient matching, and temporal consistency:
\begin{equation}
\begin{aligned}
\mathcal{L}_{\mathrm{depth}}
=
\frac{1}{N}\left\|\hat{\mathbf{D}}'-\mathbf{D}'\right\|_1
&+\frac{\alpha}{N}
\left\|\nabla_{xy}\hat{\mathbf{D}}'-\nabla_{xy}\mathbf{D}'\right\|_1
&+\frac{\gamma}{N}
\left\|\nabla_t\hat{\mathbf{D}}'-\nabla_t\mathbf{D}'\right\|_1
\end{aligned}
\end{equation}
Here $\nabla_{xy}$ and $\nabla_t$ denote spatial and temporal depth gradients,
$N$ denotes the number of video pixels, and
$\alpha,\gamma$ balance edge preservation and temporal consistency.
This transformation is used for depth supervision only; projective warping
uses the original pseudo depth $\mathbf{D}$ for geometric back-projection.

\noindent\textbf{Motion Correspondence Grounding Loss.}
Let $\mathbf{F}$ be the target flow in pixel units at resolution $H\times W$.
We scale the displacement as
$\mathbf{F}'=\mathbf{F}/\sqrt{\max(H,W)}$ and denote the corresponding prediction
by $\hat{\mathbf{F}}'$. Following RAFT~\citep{teed2020raft}, we use an $L_1$ regression loss for flow
grounding:
\begin{equation}
\mathcal{L}_{\mathrm{flow}}
=
\frac{1}{N}
\left\|
\hat{\mathbf{F}}'-\mathbf{F}'
\right\|_1
\end{equation}
When backward flow is available, the same loss is applied in the reverse
direction.

\noindent\textbf{Projective Warping Grounding Loss.}
For each supervised pair, a projection-validity mask $\mathbf{M}_{i\rightarrow j}$
is defined on the RGB reconstruction grid. A target pixel is retained when its
back-projected point has positive depth after transformation into the source
camera and projects within the source image:
$0\leq u\leq W-1$ and $0\leq v\leq H-1$.
For the set of supervised frame pairs $\mathcal{P}$, the loss is
\begin{equation}
\mathcal{L}_{\mathrm{warp}}
=
\frac{1}{N_w}
\sum_{(i,j)\in\mathcal{P}}
\left\|\mathbf{M}_{i\rightarrow j}\odot
(\hat{I}^{\mathrm{warp}}_{i\rightarrow j}-I^{\mathrm{ref}}_{i\rightarrow j})\right\|_{1}
\end{equation}

\noindent\textbf{Loss scaling and joint optimization.}
The dense supervision signals differ substantially in scale and physical units,  
including RGB intensities, depth values, and pixel displacements.  
Before combining their losses, we apply task-specific normalization and average  
each objective over its valid supervised pixels, as summarized in  
\Cref{tab:loss_normalization}. Specifically, RGB values are normalized to a  
bounded range, optical flow is scaled by image resolution, and depth is  
log-transformed to reduce its dynamic range. These normalizations bring the  
dense objectives to comparable numerical scales of approximately $O(1)$,  
without explicitly balancing their gradient magnitudes.

\begin{table}[htbp]
\centering
\caption{Formulation and normalization of the dense training objectives.
Normalization is applied before the loss weights in
\Cref{tab:appendix_loss_weights}.}
\label{tab:loss_normalization}
\small
\renewcommand{\arraystretch}{1.2}
\setlength{\tabcolsep}{5pt}
\begin{tabular}{@{}p{0.12\linewidth}p{0.25\linewidth}p{\dimexpr0.63\linewidth-4\tabcolsep\relax}@{}}
\toprule
\textbf{Objective} & \textbf{Formulation} & \textbf{Scale and normalization} \tabularnewline
\midrule
$\mathcal{L}_{\mathrm{rec}}$ & \raggedright $L_1$ reconstruction &
\raggedright Normalize RGB values to $[-1,1]$ and average the reconstruction error over video pixels. \tabularnewline
\addlinespace[2pt]
$\mathcal{L}_{\mathrm{warp}}$ & \raggedright Masked $L_1$ &
\raggedright Use RGB values in $[-1,1]$ for reconstruction and the warped-source reference;
sum errors over valid pixels and divide by their count $N_w$. \tabularnewline
\addlinespace[2pt]
$\mathcal{L}_{\mathrm{flow}}$ & \raggedright $L_1$ flow regression &
\raggedright Scale pixel displacements as $\mathbf{F}'=\mathbf{F}/\sqrt{\max(H,W)}$;
average regression errors in this domain over supervised pixels. \tabularnewline
\addlinespace[2pt]
$\mathcal{L}_{\mathrm{depth}}$ & \raggedright $L_1$ regression with spatial and temporal gradient losses &
\raggedright Transform pseudo depth as $\mathbf{D}'=\log(1+\mathbf{D})$;
compute regression and gradient losses in this domain, each normalized by $N$. \tabularnewline
\bottomrule
\end{tabular}
\end{table}

Here $H$ and $W$ are the training image height and width, $N$ is the number
of supervised video pixels, and $N_w$ is the number of valid pixels across
the warp-supervised frame pairs. The warping term is zero when no valid pixels
are available. Depth normalization applies only to the depth loss; geometric
back-projection uses the original pseudo depth.

\noindent\textbf{Total training loss.}
The video autoencoder is optimized with RGB reconstruction, perceptual,
adversarial, KL, and geometry grounding losses. The full loss is
\begin{equation}
\begin{aligned}
\mathcal{L}
=&
\lambda_{\mathrm{rec}}\mathcal{L}_{\mathrm{rec}}
+\lambda_{\mathrm{lpips}}\mathcal{L}_{\mathrm{LPIPS}}
+\lambda_{\mathrm{gan}}\mathcal{L}_{\mathrm{3D\text{-}GAN}}
+\beta\mathcal{L}_{\mathrm{KL}}  \\
&+\lambda_d\mathcal{L}_{\mathrm{depth}}
+\lambda_f\mathcal{L}_{\mathrm{flow}}
+\lambda_w a(\mathbf{x})\mathcal{L}_{\mathrm{warp}}
\end{aligned}
\end{equation}
Here $\mathcal{L}_{\mathrm{rec}}$ is the pixel-level L1 reconstruction loss,
$\mathcal{L}_{\mathrm{LPIPS}}$ is the perceptual loss, and
$\mathcal{L}_{\mathrm{3D\text{-}GAN}}$ is a spatio-temporal adversarial loss
implemented with a 3D discriminator. The discriminator receives real and
reconstructed video clips and is enabled after a warm-up stage, so the
autoencoder first learns a stable reconstruction mapping before adversarial
training sharpens spatio-temporal details.
The activation indicator $a(\mathbf{x})$ is one for RealEstate10K clips with valid
camera poses and zero otherwise, following the supervision scope in
\Cref{sec:scalable_grounding}.

\noindent\textbf{Loss weight.}
We use unit weights for RGB reconstruction, LPIPS, and all three geometry
terms, with a GAN weight of $0.1$ and a KL weight of $10^{-8}$
(\Cref{tab:appendix_loss_weights}). The 4:1 data mixture activates warping on
the 20\% RealEstate10K portion, while RGB, depth, and flow objectives supervise
both data sources. Both VGGA and VGGA w/o Geo use the same KL weight.

\begin{table}[ht]
\centering
\caption{Loss weights and training schedule used for VGGA.}
\label{tab:appendix_loss_weights}
\small
\begin{tabular}{lc}
\toprule
\textbf{Term} & \textbf{Weight or setting} \\
\midrule
RGB reconstruction $\lambda_{\mathrm{rec}}$ & 1.0 \\
LPIPS $\lambda_{\mathrm{lpips}}$ & 1.0 \\
KL regularization $\beta$ & 1e-8 \\
3D GAN $\lambda_{\mathrm{gan}}$ & 0.1 \\
GAN start step & 60k \\
Structural depth grounding $\lambda_d$ & 1.0 \\
Motion correspondence grounding $\lambda_f$ & 1.0 \\
Projective warping grounding $\lambda_w$ & 1.0 \\
Depth spatial-gradient weight $\alpha$ & 0.5 \\
Depth temporal-gradient weight $\gamma$ & 0.5 \\
Training steps & 100K \\
Learning rate & $5\times10^{-5}$ \\
GPUs & 16 H100 \\
\bottomrule
\end{tabular}
\end{table}

We train VGGA on mixed-resolution clips, including $33\times256\times256$ and
$9\times512\times512$ videos sampled at a 2:1 ratio. Unless otherwise
specified, the same optimizer, batching strategy, and latent configuration are
used for VGGA and its reconstruction-only counterpart, so the effect of geometry
grounding can be isolated from changes in model capacity or training budget.

\subsection{Downstream Latent Generator}
For downstream video generation, each autoencoder is frozen and a Latte-XL/1
latent video generator~\citep{ma2024latte} is trained on its latents. We use
velocity prediction as the diffusion prediction target. All autoencoders are
evaluated with the same generator architecture, training data, and optimization
strategy. The latent generator is trained for 250K steps on 16 H100 GPUs, and
samples are decoded by the corresponding frozen autoencoder for evaluation.

\section{Sensitivity to Pseudo Labels}
\label{apd:sensitivity}
\label{apd:teacher_robustness}
We assess sensitivity to pseudo-label teachers by varying the depth or flow
estimator while keeping the other teacher and all remaining training settings
fixed.

\subsection{Sensitivity to the Depth Teacher}
We replace VGGT with Depth Anything V2~\citep{yang2024depth} or Video Depth
Anything~\citep{chen2025video}, keeping RAFT as the flow teacher.
As shown in \Cref{tab:depth_teacher_robustness}, FVD changes by at most $0.47$
on RealEstate10K and $1.57$ on UCF101 relative to VGGT, while PSNR differs
by at most $0.07$\,dB. These results demonstrate the robustness of VGGA to different depth teachers.

\begin{table}[htbp]
\centering
\caption{Sensitivity to the depth teacher. The flow teacher is fixed to RAFT.}
\label{tab:depth_teacher_robustness}
\small
\setlength{\tabcolsep}{8pt}
\begin{tabular}{lccc}
\toprule
\multirow{2}{*}{\textbf{Depth teacher}} &
\multicolumn{2}{c}{\textbf{FVD}$\downarrow$} &
\multirow{2}{*}{\textbf{PSNR}$\uparrow$} \\
\cmidrule(lr){2-3}
& RealEstate10K & UCF101 & \\
\midrule
VGGT (default) & 64.31 & 141.27 & 33.61 \\
Depth Anything V2 & 64.78 & 142.84 & 33.67 \\
Video Depth Anything & 64.05 & 140.22 & 33.54 \\
\bottomrule
\end{tabular}
\end{table}

\subsection{Sensitivity to the Flow Teacher}
We replace RAFT with FlowNet2~\citep{ilg2017flownet}, keeping VGGT as the
depth teacher. As shown in \Cref{tab:flow_teacher_robustness}, this replacement
changes RealEstate10K and UCF101 FVD by $-0.53$ and $+0.38$, respectively,
and decreases PSNR by $0.39$\,dB. Generation performance therefore remains
similar under the tested flow-teacher replacement.

\begin{table}[htbp]
\centering
\caption{Sensitivity to the flow teacher. The depth teacher is fixed to VGGT.}
\label{tab:flow_teacher_robustness}
\small
\setlength{\tabcolsep}{8pt}
\begin{tabular}{lccc}
\toprule
\multirow{2}{*}{\textbf{Flow teacher}} &
\multicolumn{2}{c}{\textbf{FVD}$\downarrow$} &
\multirow{2}{*}{\textbf{PSNR}$\uparrow$} \\
\cmidrule(lr){2-3}
& RealEstate10K & UCF101 & \\
\midrule
RAFT (default) & 64.31 & 141.27 & 33.61 \\
FlowNet2 & 63.78 & 141.65 & 33.22 \\
\bottomrule
\end{tabular}
\end{table}

Together, these comparisons support robustness of our VGGA to the pseudo-label
teachers.

\section{Visualizations of VGGA Latents}
\label{apd:latent_visualization}
\begin{figure}[htp]
  \centering
  \includegraphics[width=\linewidth]{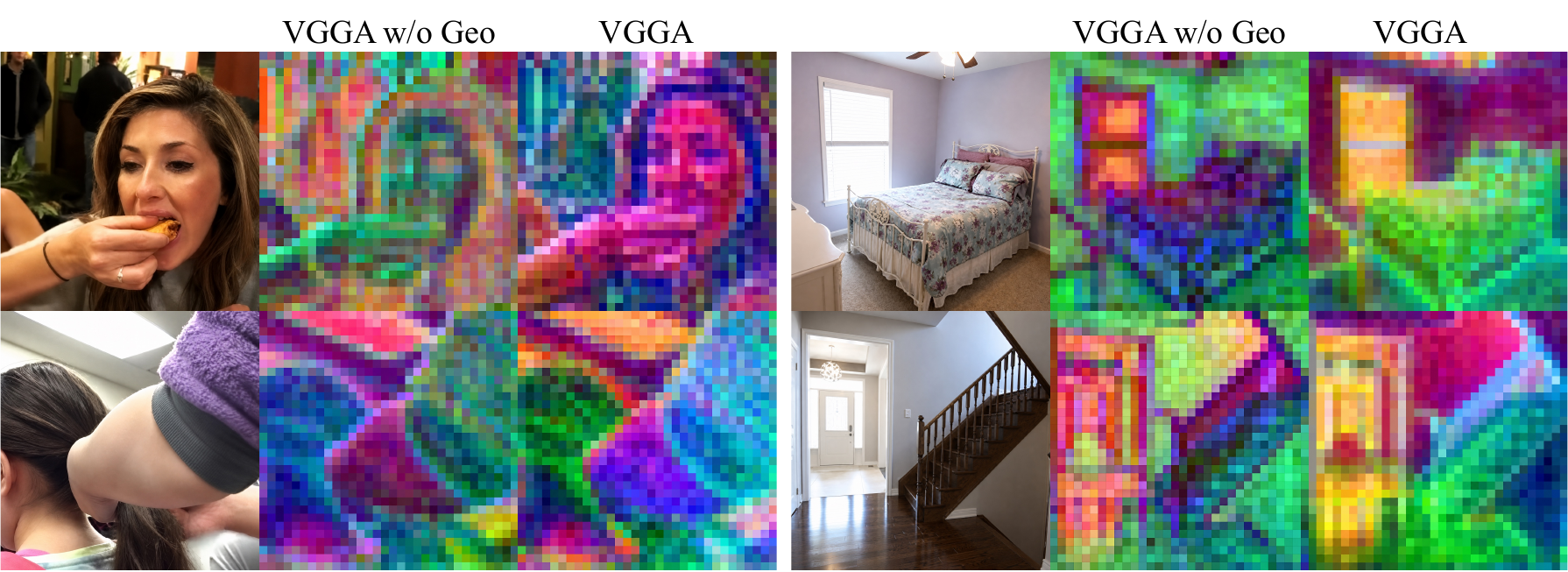}
  \vspace{-0.8em}
  \caption{\textbf{Latent Visualization.}
  Compared with the reconstruction-only baseline, VGGA exhibits a more compact
  and structured video latent space while retaining sufficient fine-grained
  visual details for faithful reconstruction.
}
\label{fig:pca_visualization}
\vspace{-0.8em}
\end{figure}

To intuitively understand the impact of geometry grounding on the latent space, we visualize the latent features using Principal Component Analysis (PCA) as shown in \Cref{fig:pca_visualization}. Compared to the reconstruction-only baseline, which often results in a scattered or entangled latent distribution, the latent space learned by VGGA exhibits a more compact and semantically organized structure. This clustering behavior suggests that our geometric constraints effectively filter out stochastic noise and redundant textural information, instead prioritizing consistent structural primitives. Despite this higher degree of abstraction, VGGA maintains sufficient expressive power to preserve fine-grained visual details, as evidenced by the high-fidelity reconstructions. These visualizations provide further evidence that geometry grounding yields a more regularized and representative latent manifold for video modeling.

\section{More Visualization Results of Video Generation}
\label{apd:vg_visualization}
\begin{figure}[ht]
\centering
\centering
  \includegraphics[width=\linewidth]{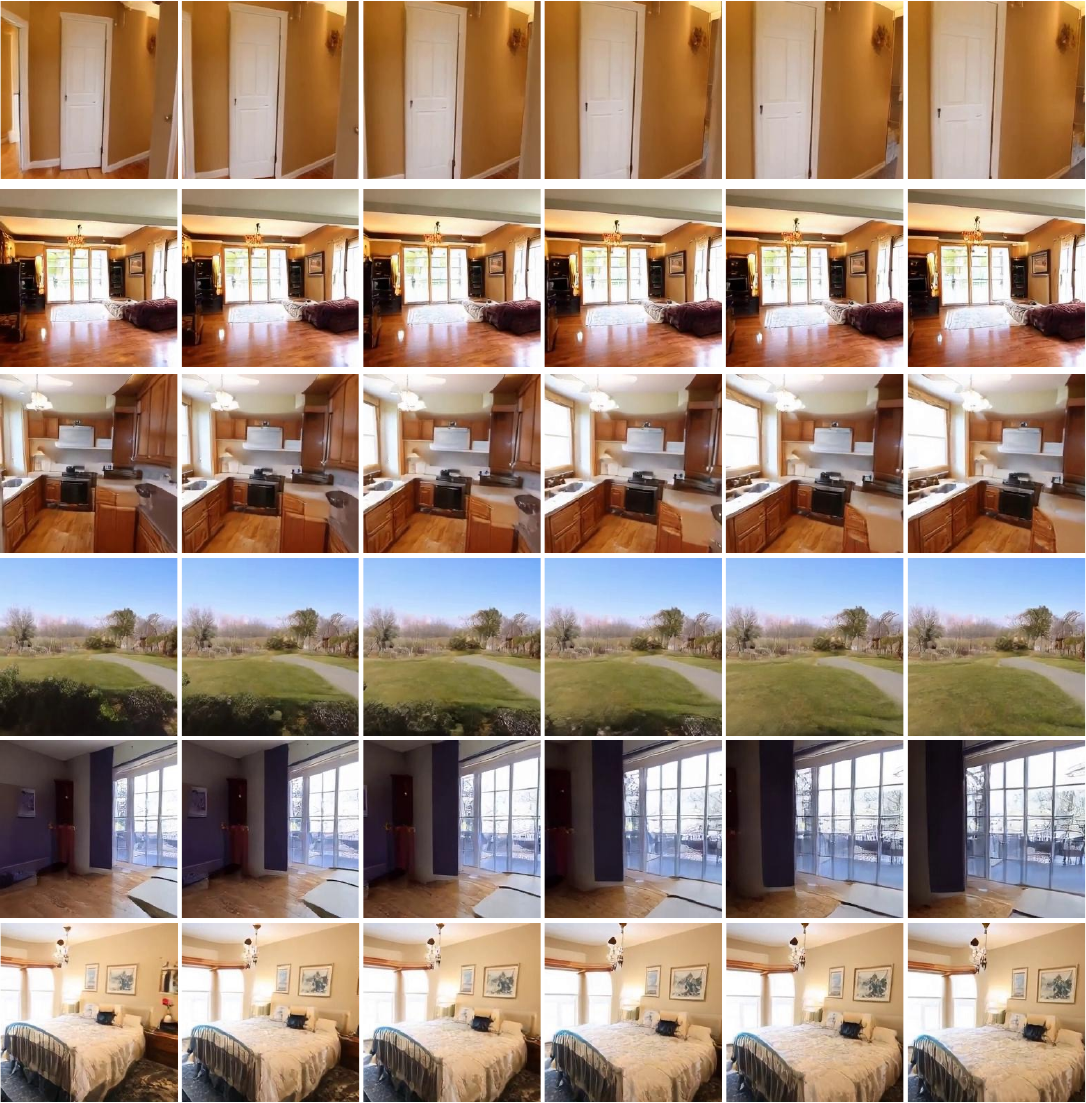}
  \vspace{-0.8em}
\caption{\textbf{Generation visualization of VGGA on RealEstate10K.}}
\label{fig:appendix_vis_realestate}
\end{figure}

\begin{figure}[ht]
\centering
\includegraphics[width=\linewidth]{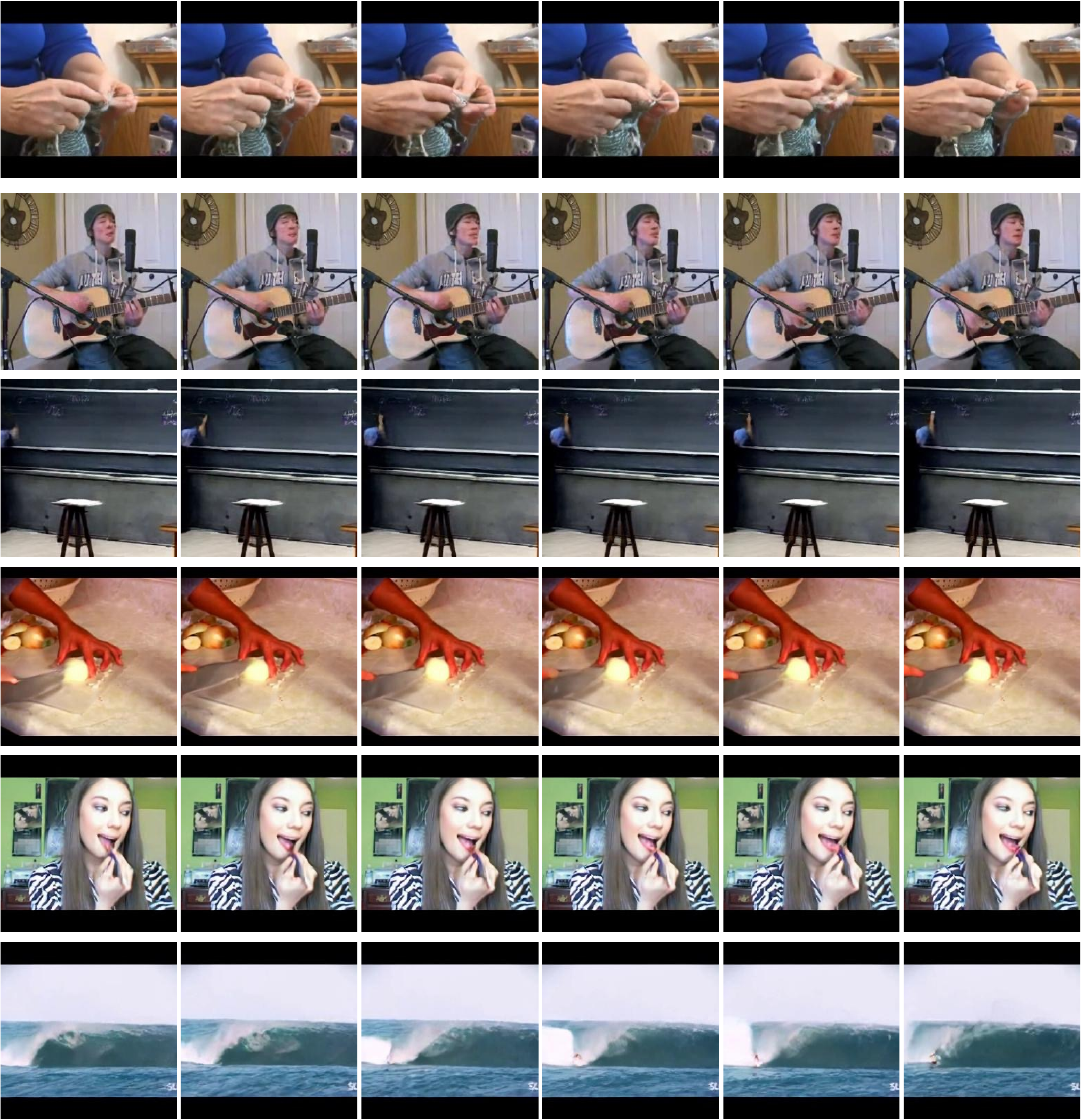}
  \vspace{-0.8em}
\caption{\textbf{Generation visualization of VGGA on UCF101.}}
\label{fig:appendix_vis_ucf101}
\end{figure}

\begin{figure}[ht]
\centering
\includegraphics[width=\linewidth]{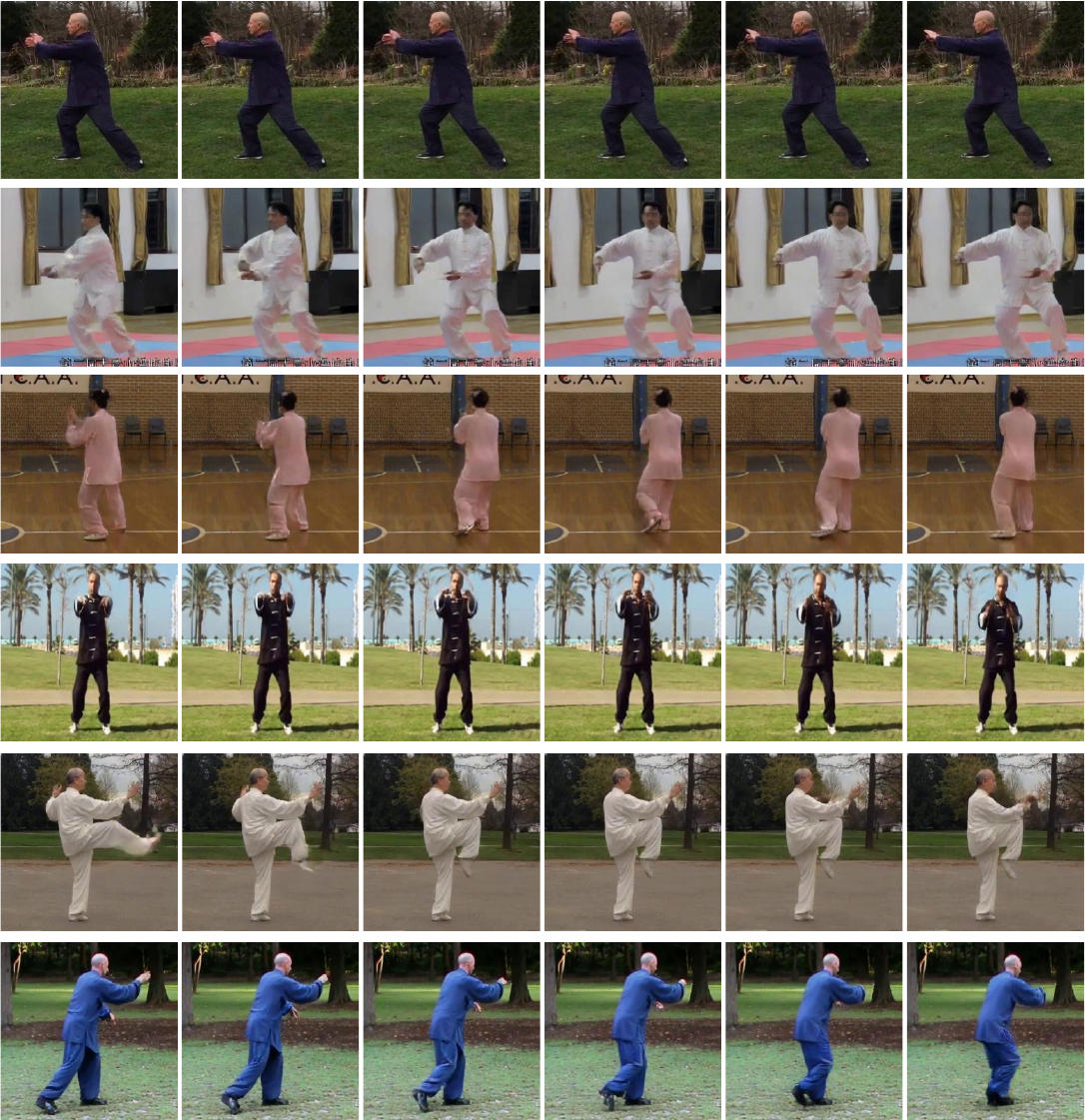}
  \vspace{-0.8em}
\caption{\textbf{Generation visualization of VGGA on Taichi-HD.}}
\label{fig:appendix_vis_taichi}
\end{figure}

We provide additional qualitative generation results on the three evaluation
datasets used in the main paper. As shown in~\Cref{fig:appendix_vis_realestate,fig:appendix_vis_ucf101,fig:appendix_vis_taichi}, these visualizations complement the quantitative
metrics by showing the effect of VGGA on camera consistency, action dynamics,
and human-motion synthesis.

\section{Geometry Grounding Evaluation}
\label{apd:gc_visualization}

\subsection{Depth and Flow Visualization}
\begin{figure}[ht]
\centering
\includegraphics[width=\linewidth]{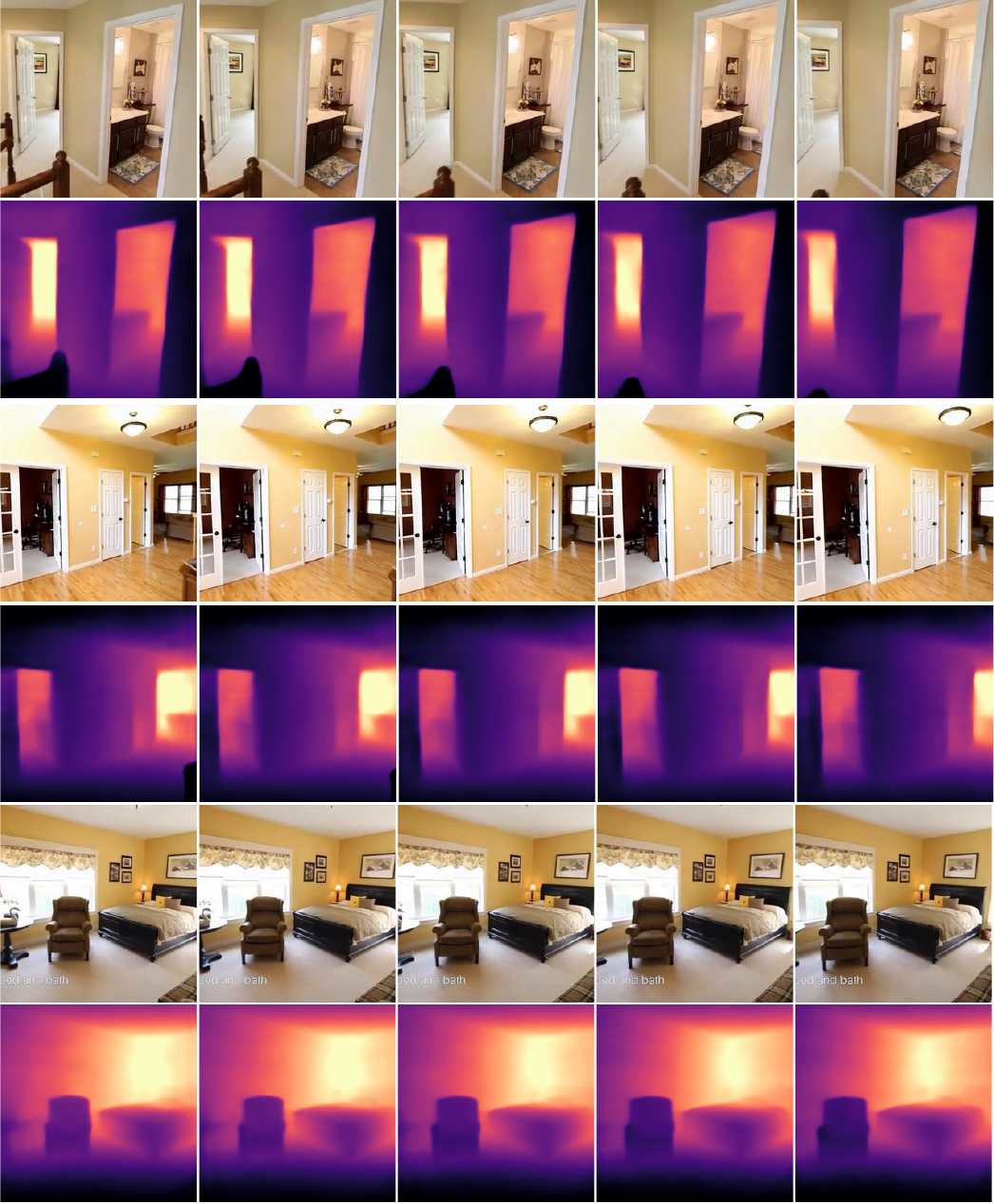}
\vspace{-0.8em}
\caption{\textbf{Visualization of structural depth grounding on dataset RealEstate10K.} VGGA predicts
temporally coherent depth maps that capture scene layout and object boundaries.}
\label{fig:appendix_depth_realestate}
\end{figure}

\begin{figure}[ht]
\centering
\includegraphics[width=\linewidth]{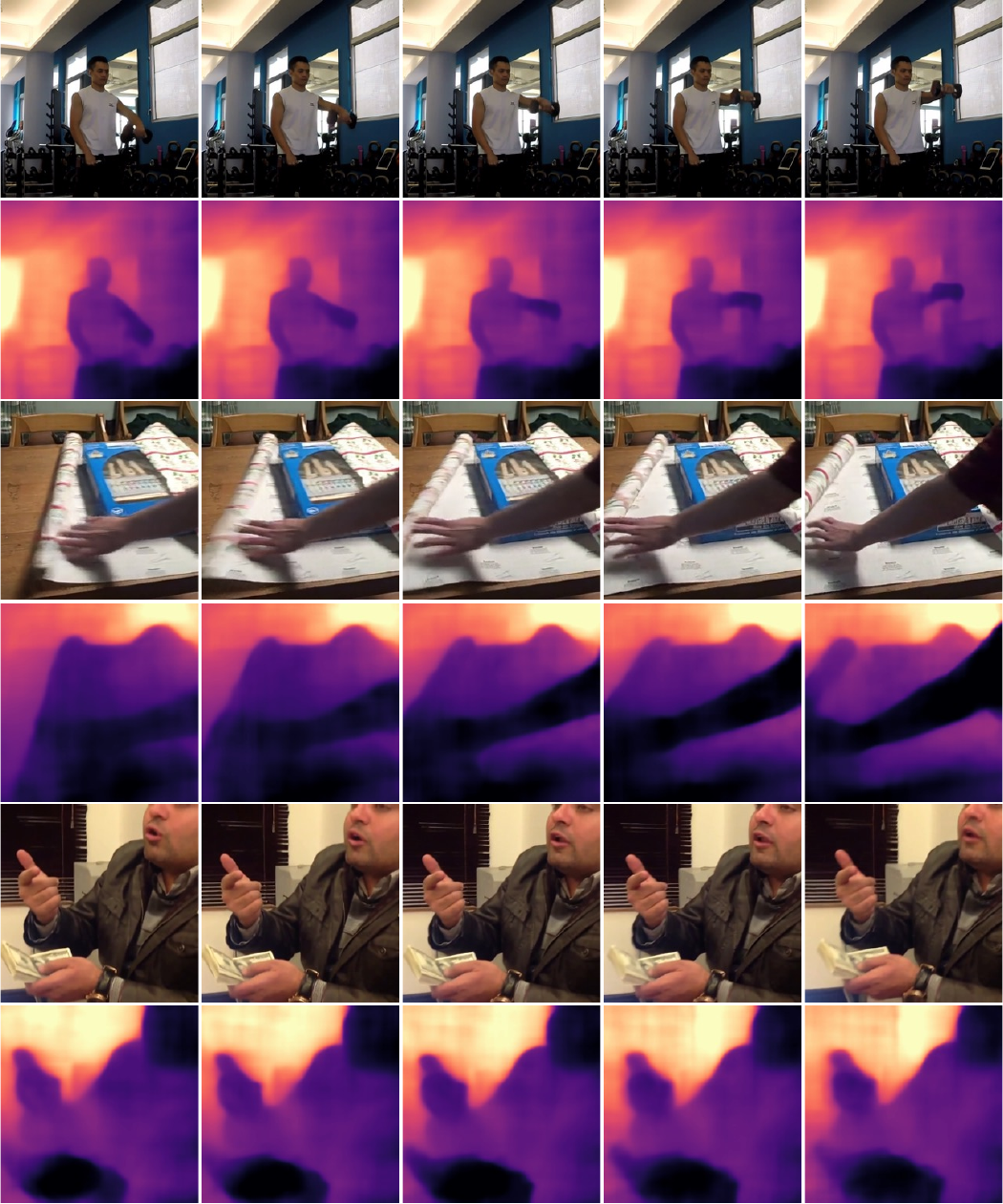}
\vspace{-0.8em}
\caption{\textbf{Visualization of structural depth grounding on dataset K400.} VGGA predicts
temporally coherent depth maps that capture scene layout and object boundaries.}
\label{fig:appendix_depth_k400}
\end{figure}

\begin{figure}[ht]
\centering
\includegraphics[width=\linewidth]{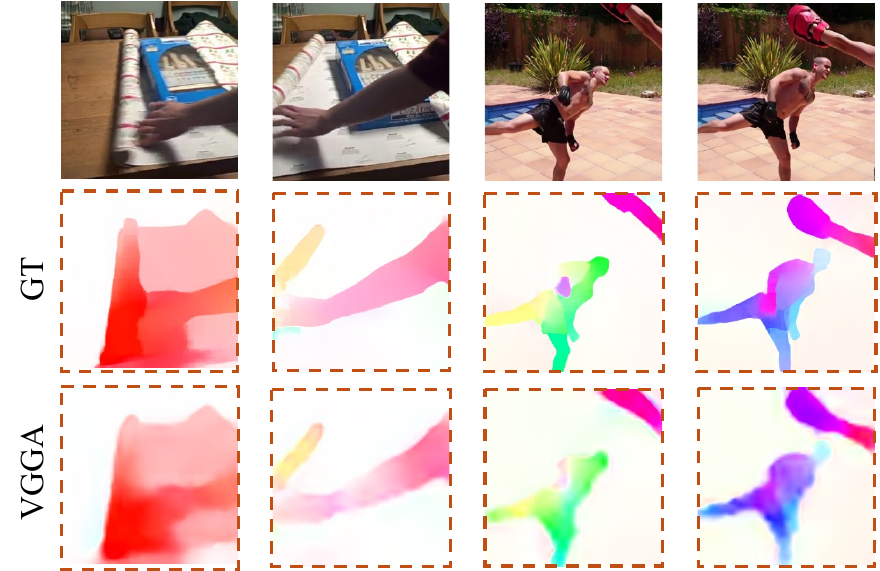}
\vspace{-0.4em}
\caption{\textbf{Visualization of motion correspondence grounding.} VGGA
predicts dense optical flow that preserves coherent object and camera motion
across frames.}
\label{fig:appendix_flow_vis}
\end{figure}

We provide additional qualitative results for the geometry
grounding branches of VGGA. These results (see~\Cref{fig:appendix_depth_k400,fig:appendix_depth_realestate,fig:appendix_flow_vis}) verify that the Video Geometry Decoder
learns meaningful depth and motion representations, rather than using the
auxiliary branches only as weak training regularizers.

\noindent\textbf{Discussion.}
The qualitative results show that VGGA recovers coherent depth and flow from
the shared decoded representation, which demonstrating that dense geometric supervision can shape video latents with
scene-layout and motion-correspondence information while remaining compatible
with RGB reconstruction.



\end{document}